\documentclass[lettersize,journal]{IEEEtran}
\usepackage{amsmath,amsfonts}
\usepackage{algorithmic}
\usepackage{algorithm}
\usepackage{array}
\usepackage[caption=false,font=normalsize,labelfont=sf,textfont=sf]{subfig}
\usepackage{textcomp}
\usepackage{stfloats}
\usepackage{url}
\usepackage{verbatim}
\usepackage{graphicx}
\usepackage{cite}
\usepackage{multirow}
\usepackage{booktabs} 
\usepackage{xcolor}
\def\ie{\emph{i.e., }}
\def\eg{\emph{e.g., }}
\begin{document}
\title{DiTalker: A Unified DiT-based Framework for High-Quality and Speaking Styles Controllable Portrait Animation}

\author{
    \IEEEauthorblockN{
        He Feng,
        Yongjia Ma,
        Donglin Di,
        Lei Fan,
        Tonghua Su, \textit{Member, IEEE}, \\
        and Xiangqian Wu, \textit{Senior Member, IEEE}
    }
    \thanks{
        *~Corresponding authors: Tonghua Su (thsu@hit.edu.cn), Lei Fan (lei.fan1@unsw.edu.au).\par
        He Feng, Tonghua Su, and Xiangqian Wu are with Harbin Institute of Technology, Harbin 150001, China (e-mail: fenghe021209@gmail.com; thsu@hit.edu.cn; xqwu@hit.edu.cn).\par
        Yongjia Ma and Donglin Di are with Li Auto, Beijing 101300, China (e-mail: maguire9993@gmail.com; didonglin@lixiang.com).\par
        Lei Fan is with University of New South Wales, Sydney 2052, Australia (e-mail: lei.fan1@unsw.edu.au).
    }
}

\maketitle

\markboth{Journal of \LaTeX\ Class Files,~Vol.~14, No.~8, August~2021}%
{Shell \MakeLowercase{\textit{et al.}}: A Sample Article Using IEEEtran.cls for IEEE Journals}

\begin{abstract}
Portrait animation aims to synthesize talking videos from a static reference face, conditioned on audio and style frame cues (\eg emotion and head poses), while ensuring precise lip synchronization and faithful reproduction of speaking styles.
Existing diffusion-based portrait animation methods primarily focus on lip synchronization or static emotion transformation, often overlooking dynamic styles such as head movements.
Moreover, most of these methods rely on a dual U-Net architecture, which preserves identity consistency but incurs additional computational overhead.
To this end, we propose DiTalker, a unified DiT-based framework for speaking style controllable portrait animation.
We design a Style-Emotion Encoding Module that employs two separate branches: a style branch extracting identity-specific style information like head poses and movements, and an emotion branch extracting identity-agnostic emotion features.
We further introduce an Audio-Style Fusion Module that decouples audio and speaking styles via two parallel cross-attention layers, using these features to guide the animation process.
To enhance the quality of results, we adopt and modify two optimization constraints: one to improve lip synchronization and the other to preserve fine-grained identity and background details.
Extensive experiments demonstrate the superiority of DiTalker in terms of lip synchronization and speaking style controllability.
Project Page: \url{https://thenameishope.github.io/DiTalker/}.
\end{abstract}


\section{Introduction}
\label{sec:intro}

\begin{figure}[!ht]
    \centering
    \includegraphics[width=\linewidth]{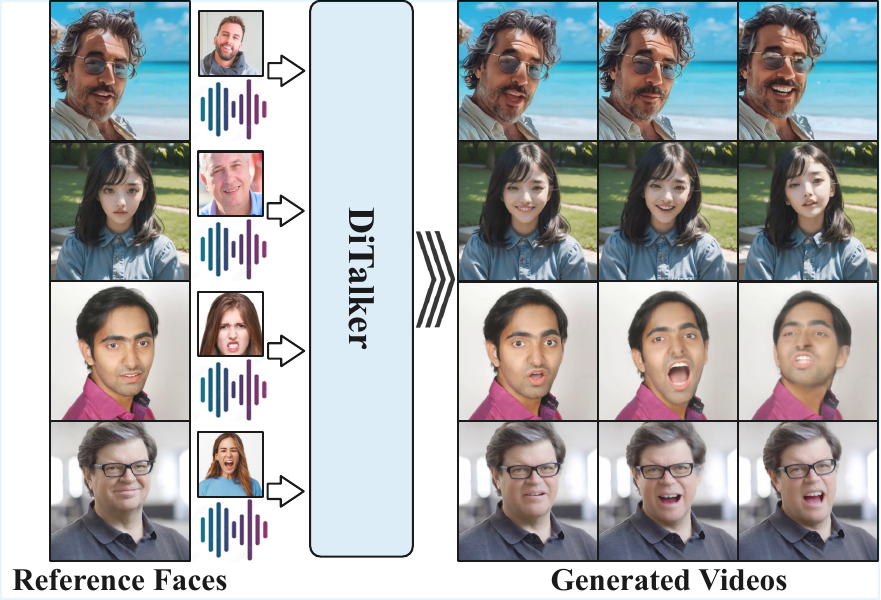}
    \caption{Given a reference face, driving audio, and a specific speaking style, DiTalker can generate high-quality talking face videos.
    The waveform on the left represents the driving audio, with the images above each audio clip showing different speaking styles.
    In the generated videos on the right, we select three temporally discontinuous frames with significant variations to demonstrate the expressiveness and vividness of the results.}
    \label{fig:teaser}
\end{figure}

\IEEEPARstart{D}{iffusion}-based video generation has achieved remarkable progress in the inherently multimodal domains of Text-to-Video (T2V) and Image-to-Video (I2V) synthesis~\cite{10845865,10909303,blattmann2023stable}.
By demonstrating the capability to produce high-quality and temporally coherent videos, these methods have stimulated widespread exploration in diverse downstream applications~\cite{10495373,yu2024wonderjourney,lin2025omnihuman1}.
Among them, human-centric video generation, notably portrait animation, has emerged as a prominent direction  \cite{10557736,11017594,10237279} due to its potential applications in video conferencing, film industry, and Virtual Reality.
Portrait animation \cite{cui2024hallo2} aims to synthesize talking face videos from a static reference face, conditioned on driving signals such as audio and style frames, as shown in Fig.~\ref{fig:teaser}.

% 引用太多了，fig 1 咋体现我们的好处，符号，抄一下别人的，style，跑配色，

%In this paper, we define speaking style as a combination of identity-specific components (\eg head poses and movements) and identity-agnostic components (\eg emotions).

In the real world, each individual has a distinct speaking style. Different people exhibit noticeable facial variations when speaking the same sentence, and the same individual may also show different facial dynamics when uttering it under different expressions. In the context of portrait animation, there are two key challenges:
(1) the precise alignment of the generated video, which involves extracting lip movement from the driving audio \cite{chung2017out}, and (2) the controllability of speaking style \cite{10413601}, which involves transferring facial expressions, head poses, and head movements.

\begin{figure}[!t]
    \centering
    \includegraphics[width=\linewidth]{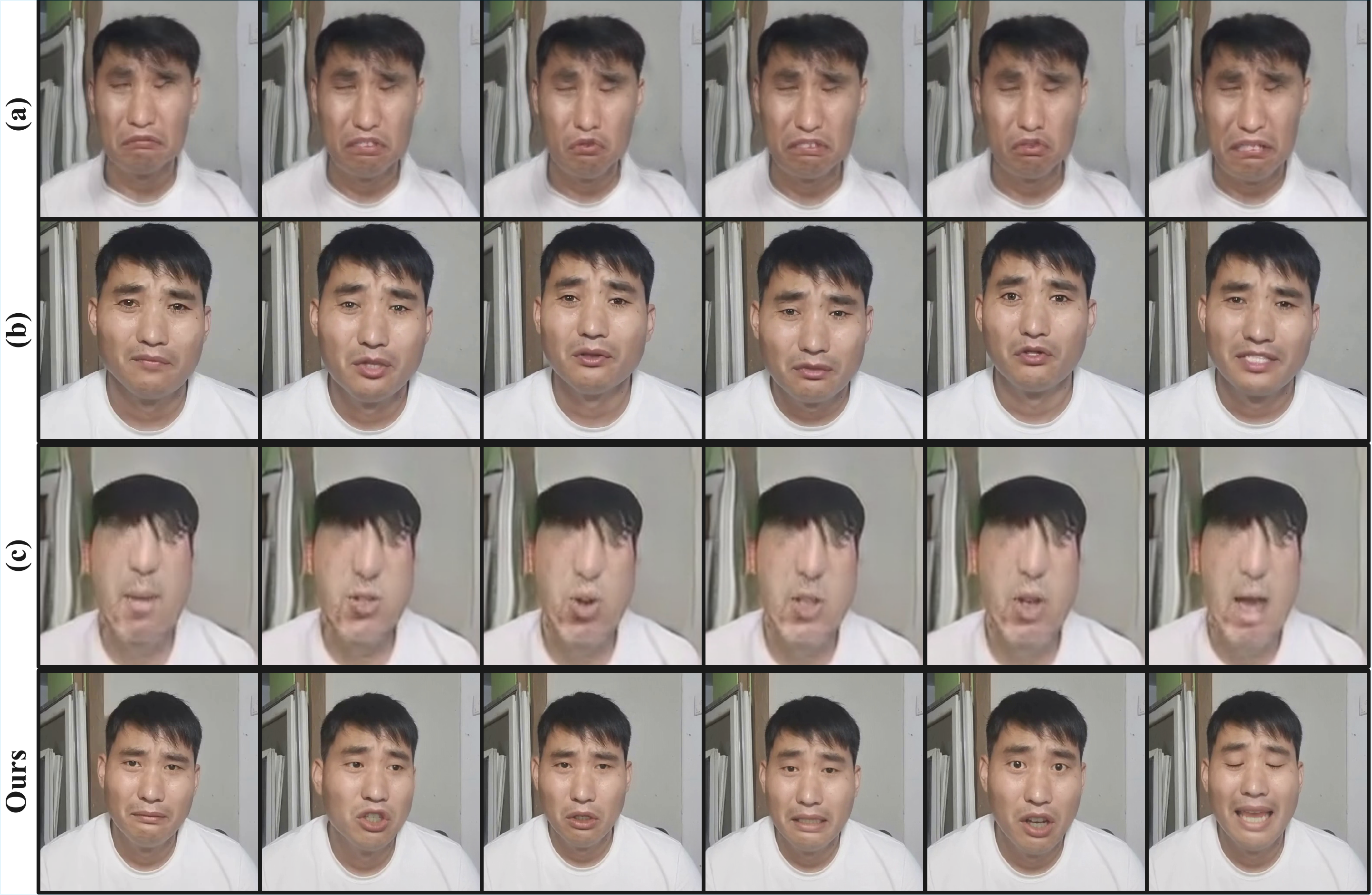}
    \caption{Comparison of our DiTalker with three categories of portrait animation methods: (a) EDTalk~\cite{tan2025edtalk}, (b) Echomimic~\cite{chen2024echomimic}, and (c) SAAS~\cite{tan2024say}. DiTalker demonstrates superior controllability of speaking style.}
    \label{fig:methods}
\end{figure}

Early studies \cite{8995571,prajwal2020lip,shen2023difftalk} have made promising progress in lip synchronization, and recent works \cite{11017594,wang2024instructavatar} have begun to explore speaking style controllable portrait animation.
For instance, InstructAvatar \cite{wang2024instructavatar} uses emotion prompts to guide emotional expression, and MoEE \cite{liu2025moee} employs emotion-specific experts for different emotions to achieve fine-grained emotional control.
SLIGO \cite{10237279} uses audio features to control both head pose and emotional expressions.
HEAD \cite{10557736} utilizes a prediction network to estimate 3DMM parameters \cite{blanz2003face} and drive facial deformation for style control.
These approaches can be categorized into three groups based on their methodology and resulting limitations: (a) those \cite{liu2025moee,wang2024instructavatar} solely guided by emotion templates or experts, which fail to capture identity-specific speaking styles; (b) those \cite{chen2024echomimic,cui2024hallo3,10237279} imprecisely predicting head pose and expressions from audio features; 
and (c) those \cite{ma2023styletalk,tan2024style2talker,10557736} relying on statistical parametric models like 3DMM, thereby limiting generalization and emotional expressiveness.
As illustrated in Fig.~\ref{fig:methods}, the comparison of inference results under the same source face, driving audio, and speaking style reveals the inherent limitations of previous methods. These limitations include facial distortions, a lack of expressive emotions, and monotonous head poses and movements, and they compromise the naturalness and vividness of the generated talking videos.

Recent diffusion-based portrait animation methods such as EMO \cite{tian2025emo} and Echomimic \cite{chen2024echomimic} employ a dual U-Net architecture, including a Reference Net that is used to extract fine-grained features of the reference source face, and a Denoising Net that is responsible for generating results.
This architectural design achieves more natural and high-quality head movements in generated talking face videos, significantly outperforming previous GAN-based approaches \cite{10557736,8995571,zhang2023sadtalker,yereal3d}.
While this design effectively ensures identity consistency, the introduction of a Reference Net incurs additional computational and training overhead. 
Moreover, current research lacks investigation into applying single-diffusion architectures, particularly Diffusion Transformer (DiT) \cite{peebles2023scalable}, which has shown superior performance over U-Net counterparts in both T2V and I2V \cite{yang2024cogvideox} generation tasks.
This gap highlights the opportunity to explore whether a single DiT-based architecture can not only reduce architectural complexity and training cost but also maintain, or even enhance, generation quality while keeping identity consistency for portrait animation.

In light of the above limitations, we propose \textbf{DiTalker}, a unified DiT-based framework for speaking style controllable portrait animation, conditioned on a static reference face, driving audio, and optional style frames.
DiTalker comprises three key components: a DiT generation backbone, a Style-Emotion Encoding Module (SEEM), and an Audio-Style Fusion Module (ASFM).
Specifically, SEEM is designed to explicitly extract speaking style features through two parallel branches.
The style branch processes style frames and phonemes extracted from driving audio to generate style embeddings, while the emotion branch encodes T5-extracted \cite{raffel2020exploring} emotional prompts from the same frames to produce emotion embeddings.
Notably, unlike prior methods like EAT~\cite{Gan_2023_ICCV}, TalkCLIP~\cite{11045546} (employing a separate emotion branch), and StyleTalk++~\cite{10413601} (employing a separate style branch), our SEEM comprises emotion and style branches. This design explicitly disentangles the head pose and movements derived from style frames from the expressed emotions, enabling independent control over each. The style branch models identity-specific speaking styles, including eye, mouth, and head movements, from style frames. Concurrently, the emotion branch models identity-agnostic emotions via an emotion template design.
This architecture ensures both the preservation of head pose from style frames and precise control over emotions.

ASFM takes audio embeddings (extracted from Whisper \cite{bain2022whisperx}) and style embeddings as inputs, disentangling audio content from speaking style through the parallel audio and style cross-attention layers.
The outputs of the two attention layers are fused via scaled element-wise addition to produce an audio-style embedding, which is then injected into the next self-attention layer of the DiT backbone.
By doing this, DiT decouples and balances audio and style attention weights, achieving accurate lip synchronization and style controllability while leveraging pre-trained weights from DiT's other layers.
Unlike the dual U-Net architecture methods \cite{wei2024aniportrait, xu2024hallo, chen2024echomimic}, we directly add noise to the reference face and input it into the DiT backbone for denoising, thereby eliminating the additional computational and memory overhead introduced by the Reference Net.

To compensate for the loss of identity and background details due to the absence of the Reference Net and to further improve lip synchronization, we adopt and modify two optimization constraints:
(1) a Latent Space Identity Loss \cite{yu2024representation}, which aligns latent representations from DiT and DINOv2 \cite{oquabd2023inov2} to ensure the preservation of identity and background details; 
(2) a Latent Space Lip Sync Loss, which aligns the mouth region of the coarse frames decoded from DiT's latent representations with SyncNet \cite{chung2017out} to enhance lip synchronization.

Overall, our contribution can be summarized as follows:
\vspace{-3pt}
\begin{itemize}
    \item We propose DiTalker, a unified DiT-based framework for speaking style controllable portrait animation, which significantly improves computational efficiency compared to other dual U-Net approaches. 
    \item We introduce a Style-Emotion Encoding Module and an Audio-Style Fusion Module that effectively decouple driving audio and speaking style, enhancing lip synchronization and allowing for precise control over styles.
    \item We modify a Latent Space Lip Sync Loss to enhance lip synchronization, which can be seamlessly integrated into various portrait animation methods.
    \item Extensive qualitative and quantitative experiments on the HDTF \cite{zhang2021flow} and CelebV-HQ \cite{zhu2022celebvhq} datasets demonstrate the superiority of DiTalker in both accurate lip synchronization and speaking style controllability.
\end{itemize}

\section{Related Work}

\noindent\textbf{Portrait Animation.}
Early methods such as Wav2Lip \cite{prajwal2020lip} and SadTalker \cite{zhang2023sadtalker}, typically employ a ``divide-and-conquer'' strategy by using separate modules to extract intermediate representations from audio and reference faces before synthesizing final talking videos.
HFWB \cite{10464342} uses a hierarchical feature warping and blending model with low- and high-level features to achieve portrait animation.
Face2Vid \cite{8995571} generates talking face videos by predicting mouth landmarks with multimodal inputs and synthesizing frames using a generation network.
Although these approaches achieve reasonable lip synchronization, they generally produce limited lip movements, monotonous head poses, and poor speaking style controllability.
Recent methods have shifted toward diffusion-based methods \cite{ho2020denoising}. Diffused Heads \cite{stypulkowski2024diffused} introduces diffusion models into the portrait animation domain by injecting audio features through modified group normalization. DiffTalk \cite{shen2023difftalk} incorporates audio conditions by concatenating them with facial keypoints, which are then used as keys and values in cross-attention layers. Subsequent methods \cite{chen2024echomimic,xu2024hallo,cui2024hallo2,tian2025emo,jiang2024loopy} adopt semantically rich audio features extracted by Wav2Vec \cite{baevski2020wav2vec} or Whisper \cite{bain2022whisperx} as inputs for cross-attention layers, and typically employ a dual U-Net architecture to maintain identity consistency and background details.

Beyond precise lip synchronization, several methods \cite{10557736,liu2025moee,10237279,gan2023efficient, ma2023styletalk,tan2023emmn, wang2024instructavatar, zhai2023talking,11017594} have explored generating style controllable talking face videos.
EAT \cite{Gan_2023_ICCV} uses a mapping network to generate emotion guidance from latent codes.
EAMM \cite{ji2022eamm} represents facial dynamics from emotional videos as motion displacements.
StyleTalk \cite{ma2023styletalk} uses a style network to extract 3DMM styles from reference videos.
Liu \textit{et al.} \cite{10557736} relied on predicting 3DMM coefficients from audio to control head pose in talking head generation.
Style2Talker \cite{tan2024style2talker} employs separate networks to simultaneously preserve art and emotion styles.
Chu \textit{et al.} \cite{11017594} used audio and identity-specific information as well as a dynamic adaptive context encoder and style adapter to control speaking style.
Sheng \textit{et al.} \cite{10237279} used audio features and a deep state space model that incorporates a variational autoencoder and normalizing flow to control emotional expressions and head poses.
EDTalk \cite{tan2025edtalk} takes video or audio inputs and employs three modules to control head pose and expression, including an audio-to-motion module.
EMNN \cite{tan2023emmn} uses emotion embeddings and lip motion to synthesize expressions and ensure consistency between lip movements and the overall facial expression. SAAS \cite{tan2024say} uses a multi-task VQ-VAE \cite{van2017neural} and a residual architecture for stylized talking head generation.
PD-FGC \cite{wang2022pdfgc} proposes one-shot talking head synthesis with disentangled lip, pose, and expression control via progressive representation and contrastive learning.
However, these methods often rely solely on either text or 3DMM for emotion control, or on reference videos for style transfer.
Over-reliance on 3DMM parameters may limit the model's generalizability, causing noticeable artifacts and flickering when the style video differs significantly from the source face in shape or emotion. Another limitation is the lack of explicit global emotion control, which makes it challenging to produce expressive results when the expression in the style video is subtle.

Unlike these methods, DiTalker incorporates a Style-Emotion Encoding Module to process these style cues.
This module extracts the corresponding style features by fusing 3DMM parameters extracted from style frames and phonemes via a transformer encoder, along with emotion features encoded by T5, and then injects them into the DiT backbone to guide the animation process.

\begin{figure*}[!htbp]
    \centering
\includegraphics[width=\textwidth]{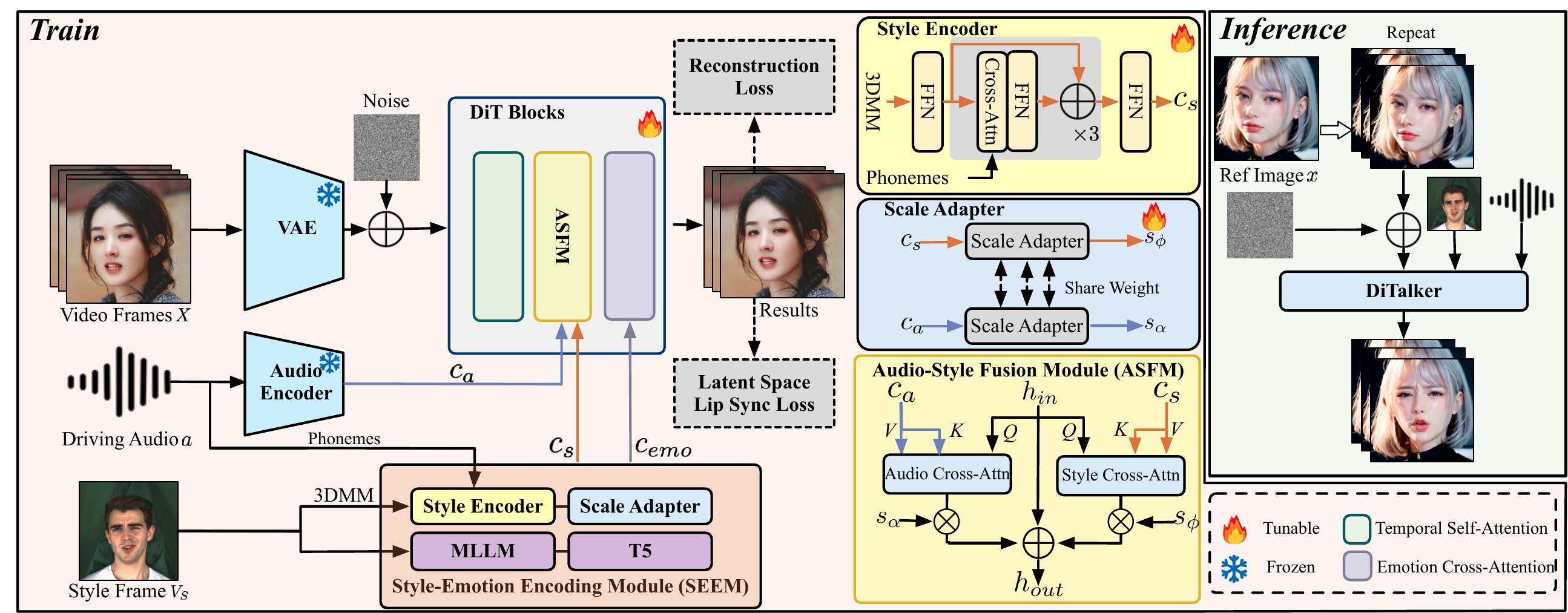}
\caption{
\textbf{Overview of our proposed DiTalker.} It consists of a DiT generation backbone, a Style-Emotion Encoding Module (SEEM), and an Audio-Style Fusion Module (ASFM).
SEEM takes style frames $V_{s}$ and phonemes (extracted from the driving audio $a$) as inputs, extracting style features $c_{s}$ and emotion features $c_{emo}$. ASFM uses $c_{s}$ and $c_{a}$ (extracted by the Audio Encoder) as inputs into the DiT backbone through two attention layers, where the outputs of the two attentions are scaled via $s_{\phi}$ and $s_{\alpha}$ extracted by the Scale Adapter in SEEM.
At inference, the DiT generation backbone animates the reference face $x$ (denoted as Ref Image) based on the features provided by SEEM and ASFM. The Emotion Cross-Attention is inserted after ASFM to enhance emotion control, where $c_{emo}$ serves as the keys and values.
MLLM denotes a multimodal large language model \cite{xu2024pllava}.
}
    \label{fig:overview}
\end{figure*}

\noindent\textbf{Diffusion in Portrait Animation.}
Early diffusion-based methods \cite{shen2023difftalk,stypulkowski2024diffused} typically employ a single-branch diffusion model for generating talking head videos.
While these approaches are effective for basic lip synchronization, they often suffer from visual artifacts and loss of details, particularly when generating large head movements.
To mitigate this, Animate Anyone \cite{hu2024animate} introduces a dual U-Net architecture, utilizing a Reference Net to extract fine-grained information from reference faces to enhance identity consistency and preserve background details.
This dual-branch strategy has since been adopted by subsequent works \cite{xu2024hallo,chen2024echomimic,cui2024hallo2,jiang2024loopy}.
More recently, some works incorporate DiT \cite{peebles2023scalable} into portrait animation.
MegActor-$\sigma$ \cite{yang2024megactor} and Hallo3 \cite{cui2024hallo3} explore its potential in talking face generation.
However, they retain the Reference Net architecture, merely replacing the original U-Net with DiT.

VASA-1 \cite{xu2024vasa} employs a single DiT architecture, but its DiT is designed to output motion parameters rather than latent representations of videos, a design choice that limits its ability to synthesize expressive talking face videos.
DiTalker uses a single DiT as the generation backbone, directly utilizing audio and style features injected via cross-attention to animate the reference face, eliminating the need for the Reference Net.

\section{Preliminary}
\noindent\textbf{Latent Diffusion Model.} 
Latent Diffusion Model (LDM) utilizes a VAE encoder \cite{rombach2021highresolution} $E$ to compress an image $I$  from the pixel space into a latent space, represented as $z_{0}=E(I)$.
During training, Gaussian noise $\epsilon$ is progressively added to $z_0$ over timesteps $t \in [1, \dots, T]$, ensuring that the final latent representation $z_{T}$ follows a standard normal distribution $\mathcal{N}(0, 1)$.
The primary training objective of LDMs is to estimate the added noise at each timestep $t$, formulated as:
\begin{equation}
    \mathcal{L}_{l d m}=\mathbb{E}_{z_{0}, t, \epsilon \sim \mathcal{N}(0,1)}\left[\left\|\epsilon-\epsilon_{\theta}\left(z_{t}, t\right)\right\|_{2}^{2}\right], \\
\end{equation}
where $\epsilon_{\theta}(\cdot)$ is the noise prediction model of the LDM.
The denoising process then learns to reverse this noise through iterative refinement, finally recovering the clean latent representation $z_0'$. A VAE decoder $D$ is used to decode it back to an image $I'$, represented as  $I'=D(z_{0}')$.
Similarly, DiTalker employs a 3D VAE \cite{xu2024easyanimate} to perform temporal downsampling of input data, achieving higher computational efficiency compared to conventional LDMs.

\noindent\textbf{Diffusion Transformer.} 
DiT shares key components with LDMs: a pair of VAEs, self-attention layers across spatial and temporal dimensions, and cross-attention layers for conditioning.
By adopting a pure transformer-based architecture, DiT improves sampling quality \cite{peebles2023scalable} and more effectively models long-range dependencies, leading to enhanced temporal consistency in generated videos \cite{yang2024cogvideox,liu2024sora}.
A core distinction in DiT is its patchification \cite{dosovitskiy2020image}, in which the latent feature $z_{0}$ is divided into a sequence of input tokens $T_k \in \mathbb{R}^{l \times s_i \times d_i}$.
Here, $s_i = hw/{p^2}$ represents the number of patches (with image height $h$ and width $w$, and the patch size $p$), $l$ is the sequence length, and $d_i$ is the dimension of each token.
Unlike U-Net-based LDMs, DiT maintains consistent feature map dimensions across layers, allowing for a long skip connection, where the input of the $j$-th DiT block is added to the output of the ($d_{b} - j - 1$)-th block, formulated as:
\begin{equation}
    z^{j}=z^{j-1} + Linear(z^{d_{b} - j - 1}),
\end{equation}
where $d_{b}$ denotes the number of DiT blocks.
In addition, DiT includes cross-attention layers, using projected hidden states as queries (Q) and conditional information like text prompts as keys (K) and values (V) for generation control.
DiTalker employs a single DiT as the generation backbone, integrating a 3D VAE \cite{xu2024easyanimate} and retaining temporal self-attention layers.

\section{Methodology}
\subsection{Overview}
Given a reference face image $x \in \mathbb{R}^{C \times 1 \times H \times W}$, driving audio
$a \in \mathbb{R}^{C_{a} \times T_{a} \times F}$, and style frames $V_{S} \in \mathbb{R}^{C \times M \times H \times W}$, our goal is to generate talking face videos $V \in \mathbb{R}^{C \times N \times H \times W}$ that achieve accurate lip synchronization and controllable speaking style.
Here, $C$, $H$, and $W$ denote the number of channels, height, and width of each frame. The variables $N$ and $M$ denote the frame lengths of the output and style frames, respectively. For the driving audio $a$, $C_a$, $T_{a}$, and $F$ represent the number of audio channels, time steps, and frequency bins.
During the generation process, $x$ provides the identity information for $V$, $a$ governs the articulation of the lip region for synchronized speech, and $V_{S}$ serves as a source of style information, guiding expression and pose variations throughout the video.

As illustrated in Fig. \ref{fig:overview}, 
DiTalker consists of three main components: a DiT backbone, a Style-Emotion Encoding Module (SEEM), and Audio-Style Fusion Modules (ASFM) that are integrated into each DiT block.
The DiT backbone processes input frames $X \in \mathbb{R}^{C \times N \times H \times W}$, driving audio $a$, and style frames $V_S$ to generate talking head videos $V$.  
The SEEM takes $V_{S}$ and the phonemes extracted from $a$ as inputs, producing style embeddings $c_{s} \in \mathbb{R}^{m_{s} \times d_{\tau}}$ and emotion embeddings $c_{emo} \in \mathbb{R}^{m_{e} \times d_{\tau}}$ to guide the DiT backbone's animation process.
The ASFM is integrated into the DiT backbone and injects the conditional features $c_{a}$ and $c_{s}$ via two parallel cross-attention layers.
To provide a facial shape prior and prevent potential facial distortion in $V$, facial keypoints $P \in \mathbb{R}^{C \times M \times H \times W}$ are extracted from the style frames using DWPose \cite{yang2023effective}. 
A lightweight 3-layer 3D CNN (referred to as Pose Adapter, which is omitted from Fig. \ref{fig:overview} for clarity) then transforms $P$ into $z_{pose} \in \mathbb{R}^{c \times n \times h \times w}$, which guides the head pose in $V$ through:
\begin{equation}
   z_{0} = z_{0} + w_{p} \cdot z_{pose},
\end{equation}
where $z_{0}\in \mathbb{R}^{c \times n \times h \times w} $ denotes the compressed representation of $X$ by $E$, and $w_{p}=0.1$ controls the influence magnitude of $z_{pose}$.

\subsection{Style-Emotion Encoding Module}
To explicitly control speaking style, we introduce the Style-Emotion Encoding Module (SEEM), which
consists of two main branches: a style branch for extracting style embeddings \(c_s\) and an emotion branch for extracting \(c_{emo}\).
This disentangled design aligns with our definition of speaking styles.
Additionally, a weight-sharing Scale Adapter \cite{liu2023stylecrafter} is employed to balance the weights of \( c_a \in \mathbb{R}^{m_a \times d_\tau} \) (encoded from audio $a$ by Whisper) and \( c_s \), generating scaling factors \( s_{\alpha}, s_\phi \in \mathbb{R}^{d_{b}} \).
In the style branch, the inputs include the style frames $V_{S}$ and phoneme labels extracted from audio $a$. 
The introduction of phonemes is inspired by \cite{ma2023styletalk}, but our approach takes it further: we simultaneously use phonemes to capture static mouth features (lip shape, mouth openness) and use Whisper features to guide the temporal dynamics between frames.
For phonemes, the extraction process involves ASR using WhisperX \cite{bain2022whisperx}, and the resulting phonemes are mapped to an index sequence through a predefined mapping table. The detailed extraction process can be found in the \emph{supplementary materials}.
We then extract 3DMM parameters $\delta_{1:M} \in \mathbb{R}^{M \times 64}$ from $V_{S}$,
which are subsequently encoded by a three-layer transformer encoder, denoted as $\psi_{E}$, to produce the hidden representations:
\begin{equation}
    H_{m} = \psi_{E}(\delta_{1:M}) = [s'_1, \dots, s'_M] \in \mathbb{R}^{M \times d_s},
\end{equation}
where $s'_{k}$ denotes the feature from 3DMM parameters $\delta_{k}$.
The phoneme labels are processed similarly using a transformer encoder and projection layers to produce $H_{p} \in \mathbb{R}^{N \times d_{s}}$.
Then, $H_{m}$ and $H_{p}$ are fed into a cross-attention layer, formulated as: 
\begin{equation}
    c_{s} = Proj\left( 
        \sigma\left(
            \frac{(H_{m} W_{Q}^{s}) (H_{p} W_{K}^{s})^{\top}}{\sqrt{d_{s}}}
        \right)
        (H_{p} W_{V}^{s})
    \right),
    \label{eq:style_embedding}
\end{equation}
where $\sigma$ denotes the Softmax function, $W_{Q}^{s}$, $ W_{K}^{s}$, and $W_{V}^{s}$ are learnable parameters, and $Proj$ denotes a linear projection followed by a reshape layer.

The emotion branch also takes $V_{S}$ as input.
A multimodal large language model \cite{xu2024pllava} is used to extract an emotion prompt $emo$ from $V_{S}$, which is then encoded using T5 \cite{raffel2020exploring} into $c_{text} \in \mathbb{R}^{m_{1} \times d_{\tau}}$. 
Here, we design a task-specific emotion prompt using the following template: ``\emph{This man/woman is [emotion] and talks}'', where \emph{[emotion]} is selected from a predefined set: \emph{\{happy, sad, angry, disgusted, surprised, fearful, neutral\}}.
To ensure identity consistency, we encode the reference face $x$ using CLIP \cite{radford2021learning} and project it into $c_{ref} \in \mathbb{R}^{m_{2} \times d_{\tau}}$.
The emotion embedding is constructed as
$c_{emo} = c_{text} \oplus c_{ref} \in \mathbb{R}^{m_e \times d_{\tau}}$, where $m_{e}=m_{1}+m_{2}$. It is used in the Emotion Cross-Attention (ECA) layer:
\begin{equation}
    \text{ECA}(z^{i}, c_{emo}) = \sigma\left(\frac{(z^{i} W_{Q}^{E})\cdot (c_{emo}W_{K}^{E})^{T}}{\sqrt{d_{\tau}}}\right)\cdot (c_{emo}W_{V}^{E}),
\end{equation}
where $z^{i}$ represents the hidden states input to the ECA, output by the ASFM, $W_{Q}^{E}$, $ W_{K}^{E}$, and $W_{V}^{E}$ are learnable parameters.
$c_{emo}$ provides global emotion control, remaining identity-agnostic and decoupling emotion from style-related factors like head pose, preventing over-reliance on 3DMM and improving stability and expressiveness in generation.

The Scale Adapter employs a three-layer Q-Former architecture \cite{zhang2024vision} to compute scaling factors $s_{\alpha}$ for $c_{a}$ and $s_{\phi}$ for $c_{s}$ through self-attention layers.
These factors are subsequently used to modulate the output features of the cross-attention layers in the ASFM.

Overall, SEEM extracts style and emotion cues from $V_{S}$ through style and emotion branches, generating style ($c_s$) and emotion ($c_{emo}$) embeddings for controllable animation.
The Scale Adapter adaptively balances audio and style contributions, enhancing speaking style control.

\begin{figure}[!t]
    \centering
    \includegraphics[width=\linewidth]{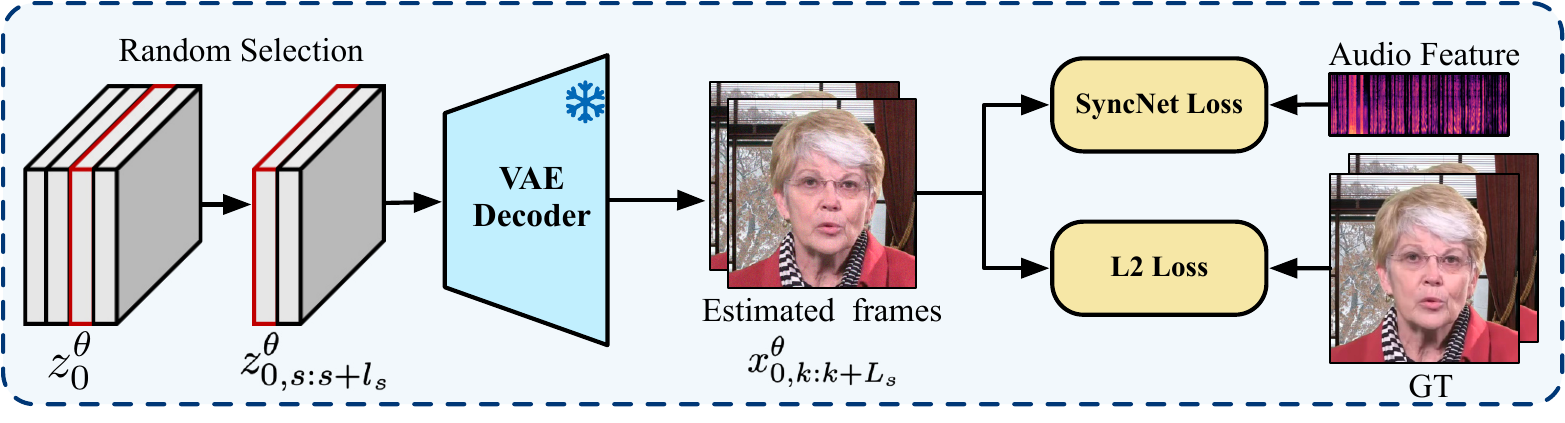}
\caption{The computation process of the Latent Space Lip Sync Loss, which includes directly estimating clean latent frames from noisy inputs, decoding them into video clips, and evaluating both the $\mathcal{L}_2$ loss and the SyncNet loss.}
    \label{fig:loss_sync}
\end{figure}

\subsection{Audio-Style Fusion Module}
The Audio-Style Fusion Module (ASFM) comprises two parts: Audio Cross-Attention (ACA) and Style Cross-Attention (SCA). These modules are designed to disentangle audio content from speaking style and process these two types of information separately, thereby achieving lip synchronization and controllable speaking style.

The ACA consists of a cross-attention layer, along with corresponding projection and normalization layers (not included in Fig. \ref{fig:overview} for clarity).
It takes hidden states $z^i$ from the previous block of the DiT backbone as queries, and audio embeddings $f_{a} \in \mathbb{R}^{N \times l \times d_{a}}$ (where $l=5$) serve as keys and values.
We fuse each frame audio embedding $f_{a}^{i}$ with its temporal context (4 preceding frames and 5 following frames) to enhance temporal consistency \cite{xu2024hallo}:
\begin{equation}
    f_{A}^{i} = (\oplus_{j=i-4}^{i+5} f_{a}^{j})  \in \mathbb{R}^{N \times L \times d_{a}},
\end{equation}
where $L=50$, $\oplus$ denotes channel-wise concatenation, and $d_{a}$ is the audio feature dimension.
The fused features are projected through a projection module  $P_{A}$ (consisting of two linear layers with $1 \times 1$ convolutions) to obtain the audio representation $c_{a} \in \mathbb{R}^{m_{a} \times d_{\tau}}$, which serves as keys and values in the ACA computation, while $z^{i}$ is used as the query after projection.

The SCA follows a similar structure to the ACA, but uses the style embedding $c_{s}$ as keys and values.
The query remains $z^i$, allowing the model to attend to style-specific signals.
After ACA and SCA are applied, we scale and fuse their outputs using scaling factors $s_\alpha$ and $s_\phi$, computed by:
\begin{equation}
\begin{aligned}
z^{i+1} &= s_{\phi}^i \cdot \sigma\left( \frac{(z^{i} W_{Q}^{S}) (c_{s} W_{K}^{S})^{\top}}{\sqrt{d_{\tau}}} \right) \cdot (c_{s} W_{V}^{S}) \\
        &\quad + s_{\alpha}^i \cdot \sigma\left( \frac{(z^{i} W_{Q}^{A}) (c_{a} W_{K}^{A})^{\top}}{\sqrt{d_{\tau}}} \right) \cdot (c_{a} W_{V}^{A}),
\end{aligned}
\end{equation}
where $s_\alpha^{i}$ and $s_\phi^{i}$ are the scaling factors corresponding to the $i$-th layer, $W_{Q}^{S}, W_{K}^{S}, W_{V}^{S}, W_{Q}^{A}, W_{K}^{A}$, and $W_{V}^{A}$ are learnable parameters.

Through this dual-attention mechanism, ASFM allows the DiT backbone to incorporate audio and style conditions independently.
The scaling factors enable dynamic weighting of each modality.
To stabilize training and mitigate the impact of randomly initialized newly added layers, we zero-initialize the output layers of the two cross-attention layers at the beginning of training.

\subsection{Loss Functions}
\noindent\textbf{Latent Space Identity Loss.} 
To enhance identity consistency and maintain fine-grained background details in $x$, we modify a Latent Space Identity Loss to align DiT representations with DINOv2.
Specifically, we randomly sample one frame of hidden states from the first $d$ layers of DiT and project it to align with the shape of DINOv2 outputs.
The alignment is enforced by minimizing the negative cosine similarity with the representation of the same frame in the pixel space, formulated as:
\begin{equation}
    \mathcal{L}_{id}(\theta,\phi):=-\mathbb{E}_{\mathbf{y}_{*},\boldsymbol{\epsilon},t}\Big[\frac{1}{d_{b}}\sum_{j=1}^{d_{b}}\mathrm{sim}(\mathbf{y}_{*}^{j},h_{\phi}({h}_{t}^{j}))\Big],
\end{equation}
where $\mathrm{sim}(\cdot,\cdot)$  denotes cosine similarity, $d_{b}$ denotes the depth of DiT, $\mathbf{y}_{*}^{j}$  denotes DINOv2 feature representation, $h_{\phi}$ denotes a linear layer, and $h_{t}^{j}$ denotes the DiT block's hidden state for the $t$-th frame at depth $j$.

\noindent\textbf{Latent Space Lip Sync Loss.} 
To enhance lip synchronization accuracy, we adopt and modify a Latent Space Lip Sync Loss  $\mathcal{L}_{\text {sync}}$.
Given a noisy latent variable $z_t$ at timestep $t$ and the noise $\epsilon_{\theta}$ predicted by the model, we directly estimate
 $z_{0}^{\theta}$, defined as:
\begin{equation}
    z_{0}^{\theta}=\frac{1}{\sqrt{\bar{\alpha}_t}} z_{t} - \sqrt{\frac{1}{{\bar{\alpha}_t}} -1}\epsilon_{\theta}.
\end{equation}
Next, we randomly sample $l_{s}$ latent frames $z_{0, s:s+l_{s}}^{\theta}$ and decode them into a coarse video $x_{0, s:s+L_{s}}^{\theta}$  using the VAE decoder $D$, where $L_{s} = l_{s} \times 4$, a factor that accounts for temporal upsampling.
We then select $L_{f}$ continuous frames $x_{0, k:k+L_{f}}^{\theta}$ and their corresponding deep speech features $c_{d, k:k+L_{f}}$, and input both of them into a pre-trained SyncNet \cite{chung2017out} to compute the SyncNet loss, where $k$ is an integer randomly sampled from the range $[s, s+L_s-L_f]$.
To ensure the quality of the reconstructed coarse frames, we apply an $\mathcal{L}_{2}$ loss.
$\mathcal{L}_{\text {sync}}$ is calculated as follows:
\begin{align}
    \mathcal{L}_{sync} = & \mathbb{E}_{z_{0, s}, x_{0, k}, t, \epsilon} \Big[ \left\|z_{0, s}^{\theta} - z_{0, s}\right\|_{2}^{2} \notag \\
    & + \lambda_{s} \cdot  \operatorname{SyncNet}\left(D(x_{0, k:k+L_{f}}^{\theta}), c_{d, k:k+L_{f}}\right) \Big],
\end{align}
where $\lambda_{s}$ is a weighting factor that balances reconstruction fidelity and synchronization accuracy.

\noindent\textbf{Overall Training Loss.} 
To encourage the model to focus on the eye region, we initially employ the LDM noise reconstruction loss with an eye mask $M_e^i \in \{0,1\}^{1\ \times h \times w}$, where $M_{e} = \bigcup_{i=1}^{4}  M_{e}^{i}, M_{e} \in \{0,1\}^{1 \times h \times w}$ represents the union of eye masks sampled from every fourth frame of a training video.
The reconstruction loss is formulated as:
\begin{equation}
\begin{aligned}
    \mathcal{L}_{rec}=& \mathbb{E}_{z_{0}, t, \epsilon \sim \mathcal{N}(0,1)}\left[\left\|\epsilon-\epsilon_{\theta}\left(z_{t}, t\right)\right\|_{2}^{2} \right. \\
    & \left. + \lambda_{eye} \cdot \left\| M_{e} \odot \epsilon- M_{e} \odot \epsilon_{\theta}\left(z_{t}, t\right)\right\|_{2}^{2}  \right],
\end{aligned}
\end{equation}
where $\lambda_{eye}$ balances the contribution of the eye-focused term, and $\odot$ denotes element-wise multiplication.
The overall loss $\mathcal{L}$ is formulated as: 
\begin{equation}
    \mathcal{L} = \mathcal{L}_{rec} + \lambda_{id} \mathcal{L}_{id}  + \lambda_{sync} \mathcal{L}_{sync},
\end{equation}
where $\lambda_{id}$ and $\lambda_{sync}$ are used to balance the two loss functions.

\section{Experiments}

\subsection{Implementation Details}
\noindent\textbf{Experimental Setups.}
Our method is based on EasyAnimate \cite{xu2024easyanimate}, a DiT-based I2V generation model.
All experiments were conducted using 8 Nvidia A800 GPUs.
We used CelebV-Text \cite{yu2023celebv} and the Hallo3 dataset as the English audio training set.
For the Chinese audio training set, we utilized DH-FaceVid-1K \cite{di2024facevid}.
We first fine-tuned the I2V weights for 10K iterations with a batch size of 1, employing the AdamW optimizer with a learning rate of 4e-6.
We then trained the ASFM and style branch of SEEM using filtered high-quality video samples from DH-FaceVid-1K, CelebV-Text, and Hallo3, with a learning rate of 2e-5.
The details about filtering rules can be found in the \emph{supplementary materials}.
When training the ECA and emotion branch of SEEM, we used the DH-FaceEmoVid-150 dataset \cite{liu2025moee}. 
We did not use the MEAD \cite{wang2020mead} dataset due to the short duration of individual videos (usually less than 2 seconds) and the uniform green screen background, which may cause potential overfitting and visual quality degradation.
To enhance the quality of generated results, all input conditions were dropped with a probability of 0.1. 
The weights $\lambda_{id}$ and $\lambda_{sync}$ were each set to 0.1, $\lambda_{eye}$ was set to 10, $\lambda_{s}$ was set to 0.5, $l_{s}$ was set to 2, $L_{s}$ was set to 8, and $L_{f}$ was set to 5.
Additionally, we observed that the pre-trained SyncNet did not perform accurately when evaluating audio in languages such as Chinese (\eg values of Sync-C and Sync-D).
To address this, we adopted \cite{li2024latentsync} and retrained a SyncNet using the DH-FaceVid-1K dataset for computing $\mathcal{L}_{sync}$ and subsequent quantitative experiments.

\begin{table*}[t]
\centering
\caption{
Quantitative comparison of our DiTalker with SOTA baseline methods on the HDTF and CelebV-HQ test sets. 
Bold numbers represent the best, while underlined numbers indicate the second-best.
}
\resizebox{\linewidth}{!}{
\begin{tabular}{l c  c c c c c  c c c c c}
\toprule[1.5pt]
\multirow{2}[2]{*}{Method} & \multirow{2}[2]{*}{Framework} &  \multicolumn{5}{c}{\textbf{HDTF}} & \multicolumn{5}{c}{\textbf{CelebV-HQ}} \\
\cmidrule[0.5pt](lr){3-7} \cmidrule[0.5pt](lr){8-12}
& & FID$\downarrow$ & FVD$\downarrow$ & LPIPS$\downarrow$ & Sync-C$\uparrow$ & Sync-D$\downarrow$
& FID$\downarrow$ & FVD$\downarrow$ & LPIPS$\downarrow$ & Sync-C$\uparrow$ & Sync-D$\downarrow$
\\
\midrule[1pt]
Wav2Lip \cite{prajwal2020lip} & \multirow{2}[2]{*}{GAN}  & 13.05 & 233.27  & 0.107 & \textbf{4.366} & \textbf{7.834}  & 
7.99 & 253.72 & 0.252 & \textbf{4.613}  & \textbf{8.729}\\
SadTalker \cite{zhang2023sadtalker} &    & 44.04 & 232.32 & 0.458 & 1.034 & 9.824 & 
 22.28 & 203.73 & 0.403 & 3.784 & 9.250\\
Real3DPortrait  \cite{yereal3d}&    & 18.78& 117.45  & 0.204 & 3.562 & 8.386 & 
13.56 & 112.53 & 0.299 & 4.093  & 8.266 \\
\midrule[1pt]
AniPortrait \cite{wei2024aniportrait} &   & 30.33 & 206.48  & 0.165 & 3.089 & 9.316 & 
14.36 & 132.12 & 0.300 & 2.485 & 9.865\\
Hallo  \cite{xu2024hallo}&  {\raisebox{-2.5pt}[0pt][0pt]{U-Net}}  & 15.63 & 99.46 & 0.105 & 3.732 & 8.365 & 
8.12 & 112.35 & 0.265 &  3.933  &  8.968\\

EchoMimic \cite{chen2024echomimic} &   & 33.42 & 198.12 & 0.424 & 2.559 & 8.947 & 
17.59 & 250.02 & 0.410 & 3.802   & 9.729 \\
Hallo2 \cite{cui2024hallo2} &   & 14.92 & 103.96 & 0.117 & 3.454 & 8.635 & 
7.43 & 84.15 & 0.270 & 3.907   & 8.751 \\

\midrule[1pt]

Hallo3 \cite{cui2024hallo3} &   & \textbf{12.18} & \textbf{90.88} & \underline{0.101} & 3.445 & 8.723 & 
\textbf{6.49} & \textbf{57.54} & \underline{0.126} & 3.887   & 8.885 \\

\textbf{Ours} & \multicolumn{1}{c}{\raisebox{4.5pt}[0pt][0pt]{DiT}} & \underline{13.03} & \underline{98.79} & \textbf{0.081} & \underline{3.823} & \underline{8.313}  & \underline{7.18} & \underline{65.16} & \textbf{0.123} & \underline{3.959} & \underline{8.732}\\
\bottomrule[1.5pt]
\end{tabular}
}
\label{tab:quantitative}
\end{table*}

\noindent\textbf{Test Set Preparation.}
We used HDTF \cite{zhang2021flow} and CelebV-HQ \cite{zhu2022celebvhq} as test sets for evaluating visual quality and lip synchronization.
For evaluating speaking style controllability, we selected a non-overlapping portion of the DH-FaceVid-1K and DH-FaceEmoVid-150 datasets containing 7 emotions (happy, sad, angry, disgusted, surprised, fearful, neutral).
DH-FaceVid-1K serves as the ``neutral'' data while DH-FaceEmoVid-150 provides the ``emotional'' data, forming our ``Mix Emotion'' test set.
The proportion of these 7 emotions in the Mix Emotion test set was 6:1:1:1:1:1:1.
The HDTF and CelebV-HQ test sets are English audio test data, while Mix Emotion serves as the non-English (\eg Chinese) audio test data.
All test sets were segmented into 5-second clips, with an audio sampling rate of 16 kHz, and resized to $512\times512$ at 25 fps.
For each of the three test sets, we sampled 2048 videos, totaling 6144 videos across all sets.
For each video, 4 frames were evenly sampled, resulting in a total of 24576 images.
This procedure followed prior work \cite{yu2023celebv} to ensure the reliability of FID and FVD.

\noindent\textbf{Baseline Methods.}
For GAN-based methods, we compared our approach with Wav2Lip \cite{prajwal2020lip} (MM'20), SadTalker (CVPR'23) \cite{zhang2023sadtalker}, and Real3DPortrait (ICLR'24).
For diffusion-based methods, we compared our approach with Aniportrait \cite{wei2024aniportrait}, EchoMimic (AAAI'25) \cite{chen2024echomimic}, Hallo \cite{xu2024hallo}, Hallo2 (ICLR'25) \cite{cui2024hallo2}, and Hallo3 (CVPR'25).
Notably, all other diffusion-based methods use U-Net as the generation backbone, while Hallo3 adopts a dual DiT architecture.
We excluded VASA-1 and MegActor-$\sigma$ from comparison due to the lack of official implementations.
% We also cannot directly borrow their results because the test set divisions might differ, which could introduce potential bias into the evaluation results.
In the Mix Emotion dataset, we additionally included EAMM (Siggraph'22) \cite{ji2022eamm} and EAT (ICCV'23) \cite{gan2023efficient} as baseline methods because they can explicitly control the speaking styles.
Furthermore, we conducted quantitative comparisons with single-branch baseline methods (\ie methods adopting either the style or emotion branch). These include StyleTalk (AAAI'23), EDTalk (ECCV'24), PD-FGC (CVPR'23), SAAS (AAAI'24), and TalkLip (CVPR'23). Experiments were performed on the HDTF and Mix Emotion test sets.

\begin{table}[t]
\small
\setlength{\tabcolsep}{3pt} % 缩小列间空白
\centering
\caption{Comparison of Hallo2 \cite{cui2024hallo2}, Hallo3 \cite{cui2024hallo3}, and DiTalker in terms of model parameters and inference speed for generating a 64-frame video.}
\begin{tabular}{l c c c c c}
\toprule[1.5pt]
Method & Ref Net & Backbone & Time $\downarrow$ & Frames/sec $\uparrow$ & Param. \\
\midrule
Hallo2 \cite{cui2024hallo2} & $\checkmark$ & U-Net & 100s & 0.64 & 2.42B \\
Hallo3 \cite{cui2024hallo3} & $\checkmark$ & DiT & 427s & 0.15 & 19.52B \\
\textbf{Ours} & & DiT & \textbf{40s} & \textbf{1.6} & \textbf{1.95B} \\
\bottomrule[1.5pt]
\end{tabular}
\vspace{-0.5em} % 压缩表格底部空白
\label{tab:comparison}
\end{table}

\noindent\textbf{Evaluation Metrics.}
We focused on evaluating five aspects of these methods: visual quality (VQ), temporal consistency (TC), lip synchronization (LS), emotional expressiveness (EX), and the naturalness of head movements (HM), with EX and HM relating to speaking style.
For VQ, we used LPIPS and FID \cite{heusel2017gans} to measure the single frame-level differences between the self-driven generated face and ground truth (GT).
For TC, we used FVD \cite{wang2018video} to evaluate the frame-level temporal discrepancy between the generated video and the ground truth.
For LS, we used Sync-C to evaluate the consistency between the driving audio and lip movements, and Sync-D to assess the distance.
For EX and HM, we used AKD \cite{siarohin2019animating} and F-LMD (Landmark Distance on the whole face) to evaluate style information such as expressiveness and head pose.

\subsection{Quantitative Results}
As shown in Table \ref{tab:quantitative}, on the HDTF and CelebV-HQ test sets, DiTalker achieved the overall best or second-best image and video quality, along with competitive results in Sync-C and Sync-D. 
Wav2Lip achieved higher single-frame image quality and superior Sync-C/Sync-D scores due to its focus on synthesizing only the lip region (around 3.5\% of the image) while copying the rest of the image. However, its FVD was significantly worse than those of other methods, limiting its practical applications.
Previous works \cite{11017594,zhang2023sadtalker,tan2025edtalk} also reported lower Sync-C scores compared to Wav2Lip.

\begin{table}[!t]
\centering
\caption{
Quantitative comparison between our method and single-branch baseline methods for controlling speaking style on the HDTF test set.
}
\resizebox{\linewidth}{!}{
\begin{tabular}{l c c c c c}
\toprule[1.5pt]
Method & FID$\downarrow$ & FVD$\downarrow$ & LPIPS$\downarrow$ & Sync-C$\uparrow$ & Sync-D$\downarrow$ \\
\midrule[1pt]
StyleTalk \cite{ma2023styletalk} & 33.09 & 341.16 & 0.366 & 1.910 & 11.427 \\
EDTalk \cite{tan2025edtalk} & 27.05 & 119.38 & 0.249 & 2.272 & 11.070 \\
PD-FGC \cite{wang2022pdfgc} & 48.32 & 414.86 & 0.578 & 1.964 & 9.264 \\
SAAS \cite{tan2024say} & 67.66 & 421.16 & 0.520 & 1.543 & 10.815 \\
TalkLip \cite{wang2023seeing} & 17.95 & 106.52 & 0.119 & \textbf{3.943}  & \textbf{8.200} \\
\midrule
\textbf{Ours} & \textbf{13.03} & \textbf{98.79} & \textbf{0.081} & \underline{3.823} & \underline{8.313} \\
\bottomrule[1.5pt]
\end{tabular}
}
\label{tab:disentangle}
\vspace{-5pt}
\end{table}

Although DiTalker showed slightly lower scores compared to Hallo3 on both datasets, the marginal metric advantage of Hallo3 comes with a significantly higher cost in terms of inference time and memory usage.
As shown in Table \ref{tab:comparison}, Hallo3's inference time and parameter count were 10.68$\times$ and 10.01$\times$ those of DiTalker, respectively.
Compared to methods that separately model the style branch or the emotion branch, such as StyleTalk, quantitative experiments on the HDTF dataset showed that DiTalker outperforms other methods in metrics like FID, as shown in Table~\ref{tab:disentangle}.
It is only slightly behind TalkLip in Sync-C and Sync-D. This is reasonable because, like Wav2Lip, TalkLip focuses on synthesizing the lip region. Although this focus inflates the metrics, it also introduces visible boundaries near the lip, limiting its practical application.
For speaking-style control, DiTalker also significantly outperformed baseline methods on the Mix Emotion dataset (primarily Asian faces with Chinese audio), achieving 0.103 LPIPS, 5.38 FID, 65.16 FVD, and 9.46 AKD, as shown in Table~\ref{tab:style}. Although its F-LMD was slightly worse than TalkLip's, this is consistent with TalkLip’s emphasis on the lip region, which leads to similar limitations in real-world use.

DiTalker's superior performance can be attributed to our design of SEEM and ASFM. SEEM extracts style and emotion embeddings from the style frames, which are then injected into the DiT backbone through ASFM and ECA via cross-attentions. This process explicitly guides information such as expressive emotions, lip and eye movements, head poses, and movements that reflect an identity-specific speaking style.

\begin{figure*}[htbp!]
\centering
\includegraphics[width=\textwidth]{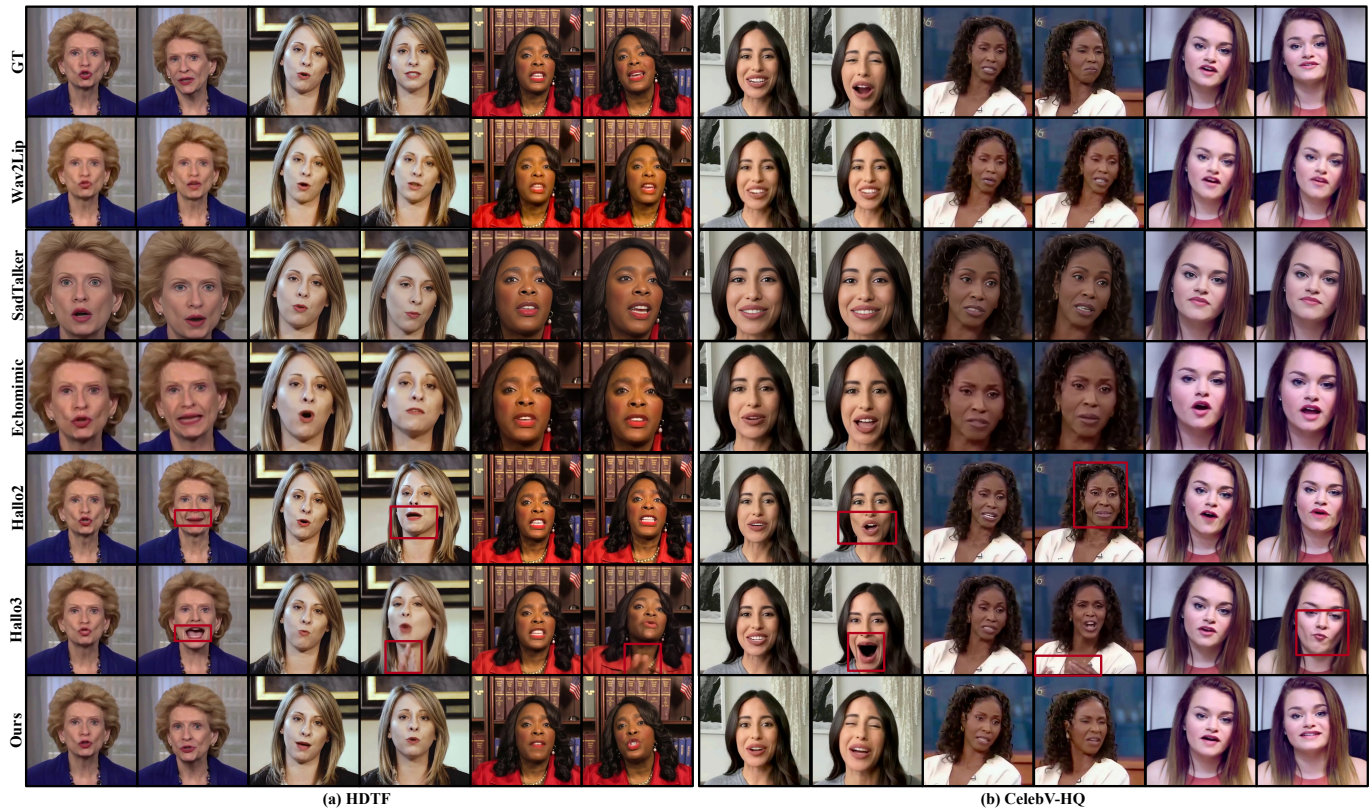}
\caption{Qualitative comparison on HDTF and CelebV-HQ test sets. DiTalker outperforms baseline methods in visual fidelity, particularly in challenging regions such as teeth and hair, while achieving comparable lip synchronization. Notably, it is the only method capable of speaking style controllable portrait animation. Please zoom in for detailed visual comparisons.}
\label{fig:quality-hdtf}
\end{figure*} 

\begin{figure}[t]
    \centering
\includegraphics[width=.98\columnwidth]{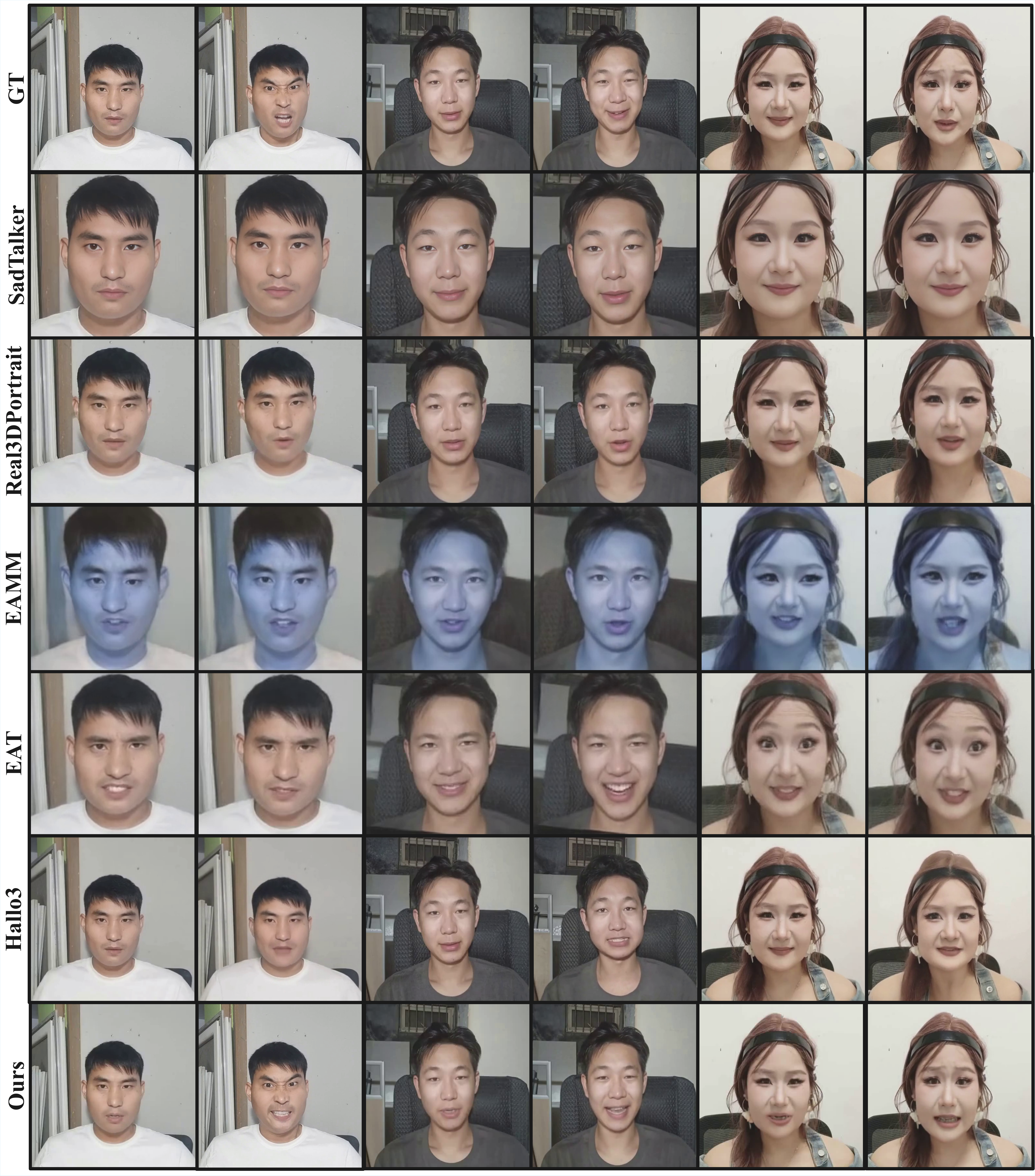}
    \caption{Qualitative comparison on the Mix Emotion test set. Compared to other baselines, DiTalker exhibits superior control over the speaking style, yielding more expressive emotions.}
    \label{fig:mixemo}
\vspace{-2pt}
\end{figure}

\subsection{Qualitative Results}
\noindent\textbf{Lip Synchronization.}
As illustrated in Fig. \ref{fig:quality-hdtf}, all methods demonstrated relatively accurate modeling of lip movements.
Wav2Lip only synthesized the lip region, resulting in poor TC (temporal consistency) and a lack of EX (emotional expressiveness) and HM (naturalness of head movements).
SadTalker generated overly smoothed sequences, yet similarly failed to exhibit sufficient EX and HM.
Echomimic achieved relatively balanced performance but enlarged the facial region, leading to a significant decrease in VQ metrics such as FID.
Hallo2 occasionally produced deep facial wrinkles, likely due to the use of CodeFormer \cite{zhou2022codeformer}, which degraded visual quality (VQ).
Although Hallo3 achieved the highest VQ, its training data lacked manual filtering of artifacts such as flashing hands, occasionally leading to unintended hand regions at the image boundaries and a reduction in TC.
In contrast, DiTalker consistently delivered superior performance across all metrics, achieving high VQ and TC, with the accurate conveyance of speaking style and head poses.
DiTalker integrates SEEM and ASFM to guide the DiT backbone in achieving lip synchronization and style controllability, adaptively adjusting attention weights via scaling factors to prevent excessive lip movements and filter out noise like hands from the training set, ensuring high-quality results.

\begin{table}[!t]
\centering
\caption{
Quantitative comparison of DiTalker with SOTA baseline methods on the Mix Emotion test set.
}
\resizebox{\linewidth}{!}{
\begin{tabular}{l c c c c c}
\toprule[1.5pt]
Method & FID$\downarrow$ & FVD$\downarrow$ & LPIPS$\downarrow$ & AKD$\downarrow$ & F-LMD$\downarrow$ \\
\midrule[1pt]
SadTalker \cite{zhang2023sadtalker} & 30.99 & 343.87 & 0.410 & 59.68 & 5.73\\
Real3DPortrait \cite{yereal3d} & 13.92 & 83.03 & 0.252 & 24.58 & 4.12\\
EAMM \cite{ji2022eamm} & 119.58 & 661.14 & 0.508 & 113.59 & 5.62\\
EAT \cite{gan2023efficient} & 76.52 & 281.38 & 0.436 & 17.93 & 3.15\\
StyleTalk \cite{ma2023styletalk} & 29.02 & 337.60 & 0.381 & 65.20 & 5.80 \\
EDTalk \cite{tan2025edtalk} & 25.39 & 135.02 & 0.289 & 28.55 & 3.78 \\
PD-FGC \cite{wang2022pdfgc} & 44.33 & 453.19 & 0.546 & 97.25 & 7.04 \\
SAAS \cite{tan2024say} & 67.08 & 524.90 & 0.599 & 99.73 & 6.88 \\
TalkLip \cite{wang2023seeing} & 18.21 & 185.51 & 0.104 & 10.52 & \textbf{2.95} \\
\midrule[1pt]
EchoMimic \cite{chen2024echomimic} & 19.91 & 234.83 & 0.398 & 30.81 & 3.56\\
Hallo2 \cite{cui2024hallo2} & 6.13 & 97.78 & 0.126 & 13.84 & 3.08\\
\midrule[1pt]
Hallo3 \cite{cui2024hallo3} & 9.25 & 107.37 & 0.173 & 20.33 & 3.21\\
\textbf{Ours} & \textbf{5.38} & \textbf{65.16} & \textbf{0.103} & \textbf{9.46} & \underline{3.01} \\
\bottomrule[1.5pt]
\end{tabular}
}
\label{tab:style}
\vspace{-5pt}
\end{table}

\noindent\textbf{Speaking Style Control.}
As shown in Fig.~\ref{fig:mixemo}, DiTalker demonstrated the capability to synthesize talking face videos with precise speaking style control, exhibiting expressive emotions (angry in the first two columns, happy in the middle two columns, and fearful in the last two columns) as well as exaggerated mouth, eye, and head movements.
In contrast, other methods, including Hallo2 and Hallo3, failed to achieve comparable performance, with their synthesized video styles remaining heavily influenced by the reference portrait.
As exemplified by Hallo3, the first two columns failed to synthesize angry emotional expressions due to the method's lack of explicit emotion modeling.
Although columns 3-4 demonstrated improved mouth aperture dynamics compared to Hallo2, they still exhibited inadequate alignment with the target emotional state.
This pattern persisted in columns 5-6, revealing consistent limitations in affective synthesis. The EAMM produced videos with significant chromatic distortions, a phenomenon we hypothesize stems from limited data diversity in its MEAD training set.
While EAT partially preserved emotional characteristics, its capacity to reconstruct physiologically plausible facial motion amplitudes remained constrained.
Quantitative evaluations further revealed that EAT's visual fidelity substantially underperformed relative to both DiTalker (FID: 76.52 vs. 5.38) and contemporary diffusion-based benchmarks.

\begin{figure}[!t]
\centering
\includegraphics[width=\linewidth]{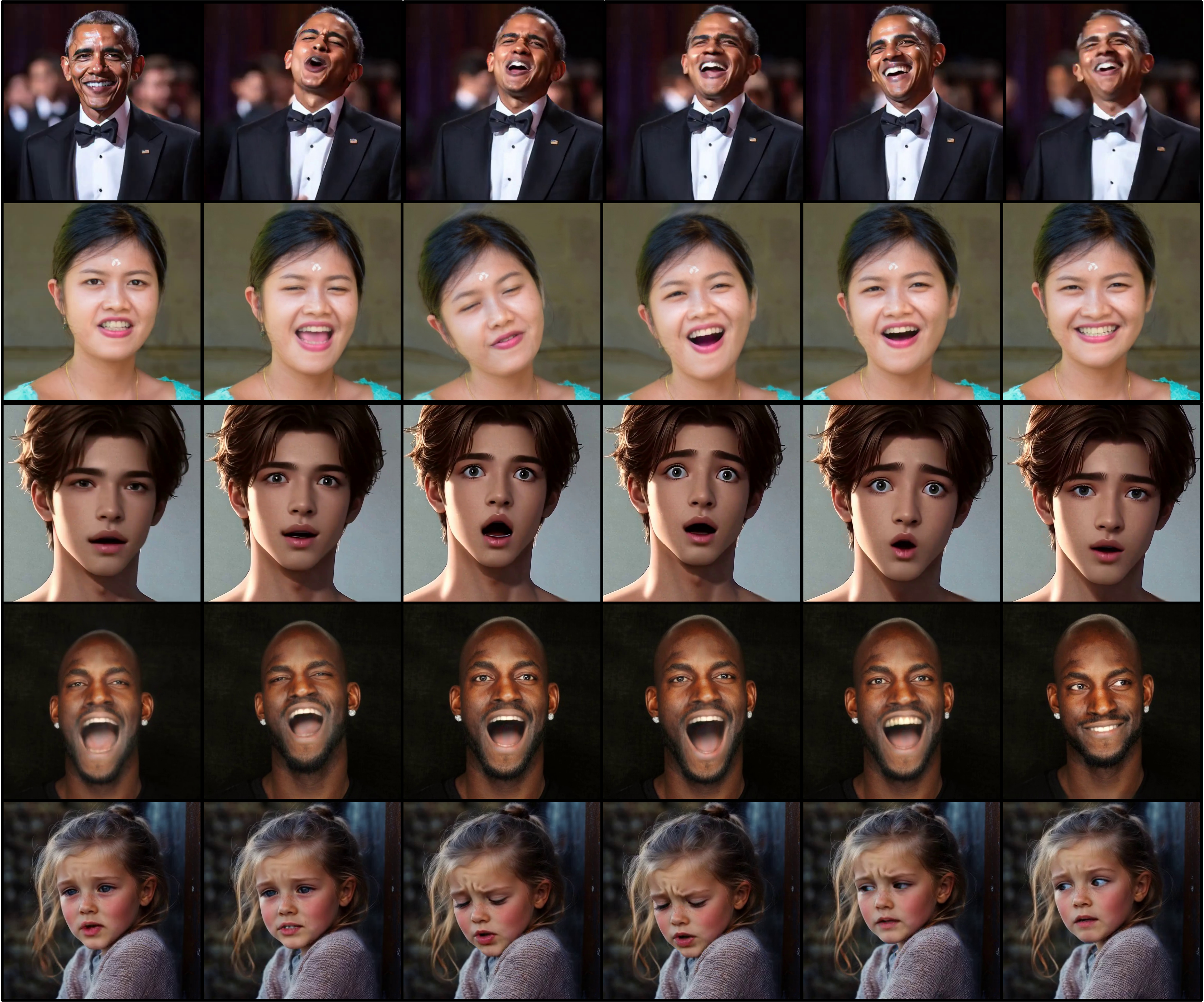}
\caption{Diverse results generated by DiTalker, including different emotions, \ie rows 1 and 4 for happy, row 2 for disgusted, row 3 for surprised, and row 5 for fearful.}
\label{fig:expressive}
\vspace{-5pt}
\end{figure}

\noindent\textbf{Challenging Conditions.}
To evaluate DiTalker’s generalization under challenging conditions, we tested it with three types of audio: noisy audio (\eg crowded streets), accented audio (\eg Indian and Scottish English accents, Cantonese, Shanghainese), and highly expressive audio (\eg shouting in anger, laughing in joy). From 100 generated videos, we selected representative examples for each case. As shown in Fig.~\ref{fig:expressive}, DiTalker maintained robust lip synchronization and accurately conveyed expressive emotion. These video results are provided on our web page.

\begin{table}[!t]
\centering
\caption{Ablation studies of newly added modules and losses.}
\resizebox{\columnwidth}{!}{
\begin{tabular}{cccc|cccc}
\toprule[1.5pt]
ASFM & $\mathcal{L}_{id}$ & $\mathcal{L}_{sync}$ & $Pose$ & FID $\downarrow$ & FVD $\downarrow$ & LPIPS $\downarrow$ & AKD $\downarrow$ \\ 
\midrule[1pt]
 & $\checkmark$ & $\checkmark$ & $\checkmark$ & 8.47 & 145.32 & 0.189 & 32.39 \\
$\checkmark$ & $\checkmark$ & $\checkmark$ & & 5.48 & 82.54 & 0.113 & 16.20 \\
$\checkmark$ & & $\checkmark$ & $\checkmark$ & 5.94 & 81.17 & 0.139 & 12.23 \\
$\checkmark$ & $\checkmark$ & & $\checkmark$ & 6.01 & 77.83 & 0.125 & 15.28 \\
\midrule[1pt]
$\checkmark$ & $\checkmark$ & $\checkmark$ & $\checkmark$ & \textbf{5.38} & \textbf{65.16} & \textbf{0.103} & \textbf{9.46} \\
\bottomrule[1.5pt]
\end{tabular}
}
\vspace{-5pt}
\label{tab:ablation}
\end{table}

\subsection{Ablation Studies}
\begin{figure}[!t]
    \centering
\includegraphics[width=\linewidth]{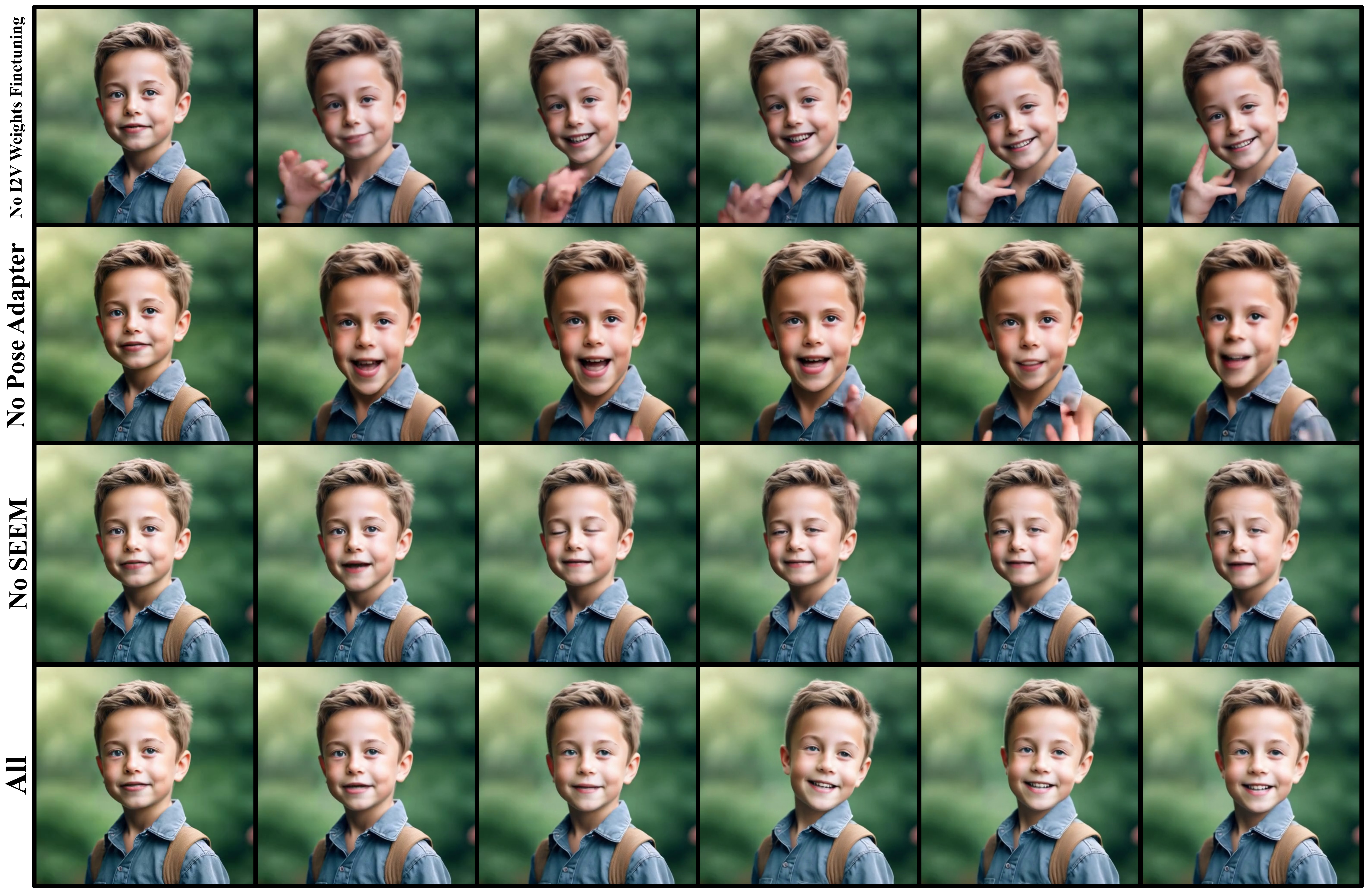}
    \caption{Ablation studies on I2V weights fine-tuning, Pose Adapter, and SEEM.
}
    \label{fig:bad}
\vspace{-5pt}
\end{figure}

\noindent\textbf{New Modules.}
We first validated the effectiveness of the newly introduced ASFM and SEEM, with or without $z_{pose}$ added to $z_{0}$, as well as $\mathcal{L}_{id}$ and $\mathcal{L}_{sync}$, through a quantitative study using the aforementioned Mix Emotion test set.
As shown in Table \ref{tab:ablation}, ASFM ensured lip synchronization and speaking style control, as reflected in the AKD (reduced by 22.93), while head pose and other speaking style-related factors further influenced the FVD (reduced by 80.16).
Similarly, adding $z_{pose}$ to $z_{0}$ (denoted as column pose in Table \ref{tab:ablation}) also affected head pose and movements, while its impact on AKD and FVD was less significant compared to ASFM (reduced by 6.74 and 17.38, respectively).
Without the explicit guidance of $z_{pose}$, the generated face sometimes deformed when making exaggerated expressions like laughing, as shown in row 2 of Fig. \ref{fig:bad}.
The introduction of $\mathcal{L}_{sync}$ improved lip synchronization, as evidenced by the improvement in the AKD (reduced by 5.82).
Meanwhile, the introduction of $\mathcal{L}_{id}$ enhanced identity consistency and the fine-grained background details from the reference face, leading to improvements across all three metrics.
In short, ASFM, SEEM, and loss functions all contribute to improved quality and fidelity in generated talking face videos.

\begin{figure}[t]
    \centering
    \includegraphics[width=\linewidth]{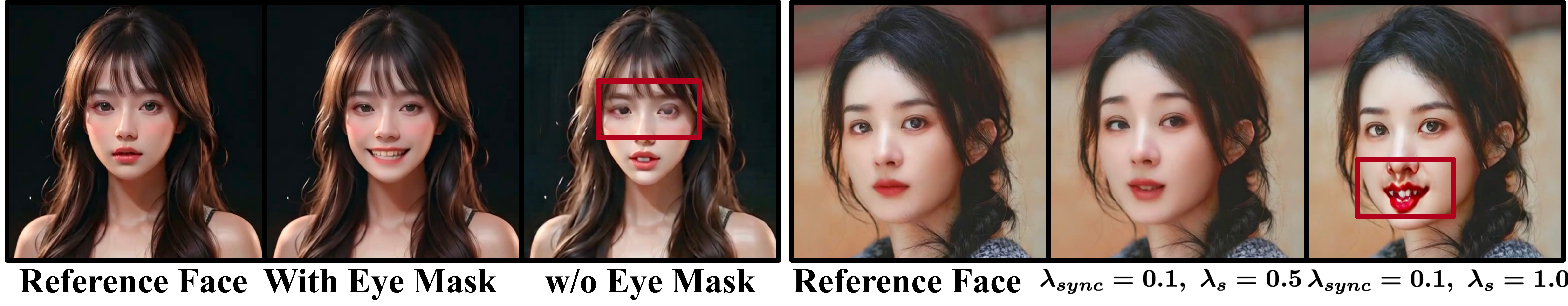}
    \caption{Ablation studies of the effect of eye mask,  $\lambda_{sync}$, and $\lambda_{s}$.}
    \label{fig:weights}
\end{figure}

\begin{table}[!t]
\centering
\caption{Ablation studies of loss function weights.}
\resizebox{\columnwidth}{!}{
\begin{tabular}{cccc|cccc}
\toprule[1.5pt]
$\lambda_{id}$           & $\lambda_{s}$   & $\lambda_{sync}$  & $\lambda_{eye}$   & FID $\downarrow$     & FVD$\downarrow$  & LPIPS $\downarrow$ & AKD$\downarrow$   \\ 
\midrule[1pt]
 0.01 & 0.5 & 0.1 & 10 &     7.23   & 69.25   & 0.117  & 9.67\\
 0.15 & 0.5 & 0.1 & 10 & 5.87   &  66.23   & 0.108   & 9.49\\
 0.1 & 0.1 & 0.1 & 10 & 5.72   &  66.17   & 0.115  & 10.01\\
 0.1 & 1.0 & 0.1 & 10 & 15.59   &  165.89   & 0.230  & 12.58\\
 0.1 & 0.5 & 0.05 & 10 & 5.57   &  71.23   & 0.112  & 13.27 \\
 0.1 & 0.5 & 0.15 & 10 & 14.21   &  127.83   & 0.209  & 15.28\\
 0.1 & 0.5 & 0.1 & 20 & 5.41   &  66.12   & 0.104  & 9.55\\
\midrule[1pt]
 0.1 & 0.5 & 0.1 & 10 & \textbf{5.38}  & \textbf{65.16}  & \textbf{0.103}  & \textbf{9.46} \\
\bottomrule[1.5pt]
\end{tabular}
}
\vspace{-5pt}
\label{tab:weights}
\end{table}

\begin{figure}[t]
    \centering
\includegraphics[width=\linewidth]{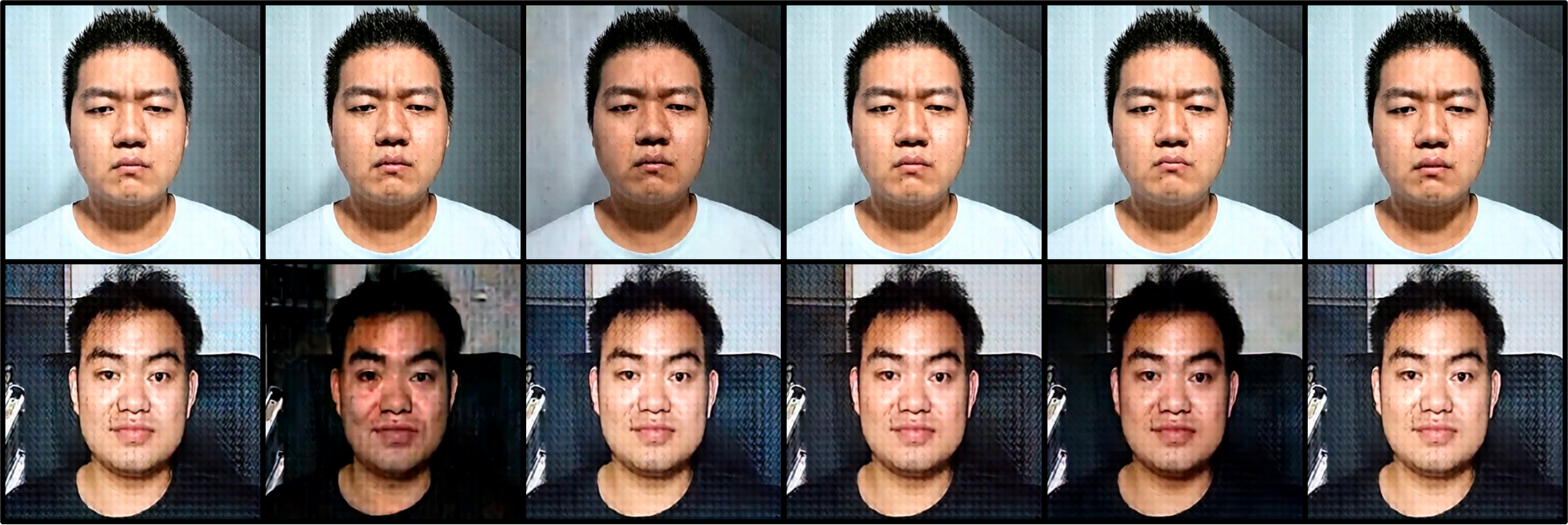}
    \caption{After EasyAnimate experienced training collapse, the generated video samples exhibited noticeable artifacts and distortions.}
    \label{fig:dead}
\end{figure}

\noindent\textbf{Fine-tuning I2V Weights.}
As discussed in the method section, our DiT backbone is for
general video generation, so we first used talking face datasets to fine-tune it before ASFM and other modules.
As shown in row 1 of Fig. \ref{fig:bad}, without this phase, directly training with the new module sometimes led to generating content outside of the human face, like random hands, due to the domain gap between general training datasets and face video datasets.
We also observed that without this training phase, the model became prone to training collapse (Fig. \ref{fig:dead}), an issue detectable only through inference results rather than loss curves.

\noindent\textbf{Loss Weights.}
We conducted an ablation study on the choice of the value for $\lambda_{id}$, $\lambda_{s}$, $\lambda_{sync}$, and $\lambda_{eye}$, as presented in Table \ref{tab:weights}.
As shown in rows 1 and 2 of Table \ref{tab:weights}, gradually increasing $\lambda_{id}$ from 0.01 to 0.1 improved FID and FVD (reduced by 1.85 and 4.09), but setting it too high (0.15) led to performance degradation.
Rows 3 and 4 regarding $\lambda_{s}$ indicated that excessively high values of $\lambda_{s}$ (as shown in Fig. \ref{fig:weights}) introduced significant artifacts in the generated lip regions and led to substantial degradation in FID, FVD, and AKD (increased by 10.21, 100.73, and 3.12).
Rows 5 and 6 demonstrated that $\lambda_{sync}$ exhibited similar behavior to $\lambda_{s}$, where excessively high values degraded visual quality greatly.
Row 7 showed that an increase of $\lambda_{eye}$ minimally affected metrics, as the eye region is small (typically under 4\%).
Based on the aforementioned studies and analysis, we selected the values of weights described in the last row.

\noindent\textbf{Disentanglement and Phoneme Input.}
We performed ablation studies on the separate style and emotion branches in SEEM as well as the phoneme input in the style branch. Specifically, we replaced their original inputs with zero tensors of the same shape to ensure normal operation.
As shown in Table~\ref{tab:disentanglement}, both the style and emotion branches contributed to lip synchronization and speaking style controllability, with the style branch playing a more significant role due to its explicit control over head pose and other speaking style-related factors. The phoneme input also contributes moderately to lip synchronization and speaking style controllability.

\begin{table}[!t]
\centering
\caption{Ablation studies of disentanglement and phoneme labels.}
\label{tab:disentanglement}
\resizebox{\columnwidth}{!}{
\begin{tabular}{cccc|cccc}
\toprule[1.5pt]
Style & Emotion & Phoneme & & FID $\downarrow$ & FVD $\downarrow$ & LPIPS $\downarrow$ & AKD $\downarrow$ \\
\midrule[1pt]
$\checkmark$ &  & $\checkmark$ & & 6.12 & 83.21 & 0.136 & 13.28 \\
 & $\checkmark$ & $\checkmark$ & & 7.64 & 128.17 & 0.157 & 24.17 \\
$\checkmark$ & $\checkmark$ &  & & 5.67 & 68.78 & 0.112 & 11.25 \\
$\checkmark$ & $\checkmark$ & $\checkmark$ & & \textbf{5.38} & \textbf{65.16} & \textbf{0.103} & \textbf{9.46} \\
\bottomrule[1.5pt]
\end{tabular}
}
\vspace{-5pt}
\end{table}

\noindent\textbf{Eye Mask.}
The use of $M_{e}$ is motivated by our observation that the eye region in generated face videos occasionally exhibits artifacts, as shown in Fig. \ref{fig:weights}.
Therefore, we employed $M_{e}$ when calculating $\mathcal{L}_{rec}$ to guide the model's attention to the eye area.
Additionally, since the eye region accounts for less than 4\% of the overall loss, we set $\lambda_{eye} = 10$ to increase its weighting in the total loss calculation.

% \vspace{-2pt}
\section{Conclusion}
In this paper, we propose DiTalker, a DiT-based framework for speaking style controllable portrait animation.
By introducing the SEEM and the ASFM, DiTalker decouples audio content from speaking styles, enabling control over lip synchronization, head poses, and emotional expressions.
Our modified Latent Space Lip Sync Loss ($\mathcal{L}_{sync}$) and adopted Latent Space Identity Loss ($\mathcal{L}_{id}$) enhance generation quality and fidelity.
Quantitative and qualitative experiments on HDTF, CelebV-HQ, and the Mix Emotion test sets show DiTalker's superior performance and computational efficiency in speaking style controllable portrait animation.

% \begin{thebibliography}{1}
\bibliographystyle{IEEEtran}
\bibliography{article.bbl}

\clearpage

\vfill

\end{document}